\documentclass[letterpaper, 10 pt, conference]{ieeeconf}  

\IEEEoverridecommandlockouts                              

\usepackage{graphicx} 
\usepackage{epsf}
\usepackage[caption=false,font=footnotesize]{subfig}
\usepackage{amsmath,amsfonts,mathtools}
\usepackage{algorithm}
\usepackage{algpseudocode}
\usepackage{multirow}
\usepackage{hyperref}

\usepackage{float}
\usepackage{stfloats}

\usepackage{xcolor}
\usepackage{comment}

\title{\LARGE \bf
Decision Making for Autonomous Driving via Augmented Adversarial Inverse Reinforcement Learning
}

\author{Pin Wang$^{1}$ \and Dapeng Liu$^{2}$ \and Jiayu Chen$^{3}$ \and Hanhan Li$^{4}$ \and Ching-Yao Chan$^{1}$%
\thanks{$^{1}$ University of California, Berkeley
{\tt\small \{pin\_wang,cychan\} @berkeley.edu}}%
\thanks{$^{2}$ Zenseact, Chalmers University of Technology
{\tt\small dapeng.liu @zenseact.com}}%
\thanks{$^{3}$ Peking University
{\tt\small 1600011060@pku.edu.cn}}%
\thanks{$^{4}$ Google Research
{\tt\small uniqueness@google.com}}%
}

\begin{document}

\maketitle
\thispagestyle{empty}
\pagestyle{empty}

\begin{abstract}
    Making decisions in complex driving environments is a challenging task for autonomous agents. Imitation learning methods have great potentials for achieving such a goal. Adversarial Inverse Reinforcement Learning (AIRL) is one of the state-of-art imitation learning methods that can learn both a behavioral policy and a reward function simultaneously, yet it is only demonstrated in simple and static environments where no interactions are introduced. In this paper, we improve and stabilize AIRL's performance by augmenting it with semantic rewards in the learning framework. Additionally, we adapt the augmented AIRL to a more practical and challenging decision-making task in a highly interactive environment in autonomous driving. The proposed method is compared with four baselines and evaluated by four performance metrics. Simulation results show that the augmented AIRL outperforms all the baseline methods, and its performance is comparable with that of the experts on all of the four metrics.
\end{abstract}


\section{Introduction}
	

The application of Reinforcement Learning (RL) in robotics has been very fruitful in recent years, ranging from flying an inverted helicopter~\cite{ng2006autonomous} to robotic manipulations~\cite{finn2016guided}\cite{singh2019end}. Though the results are encouraging, a significant barrier of applying RL to real-world problems is the required definition of the reward function, which is typically unavailable or infeasible to design in practice. Furthermore, for practical problems e.g. autonomous driving, decision making in highly interactive environments is a challenging task which requires human-like behaviours to interact with peer human drivers. 

Inverse Reinforcement Learning (IRL)~\cite{ng2000algorithms} avoids reward engineering problems by learning a reward function from expert demonstrations, and also a behavioral policy based on the reward function. In particular, Maximum entropy IRL~\cite{ziebart2008maximum}\cite{ziebart2010modeling} is commonly used, where trajectories are assumed to follow a Boltzmann distribution based on a cost function. However, because of the expensive reinforcement learning procedure in the inner loop,  this method has limited application in problems involving high-dimensional state and action spaces~\cite{ho2016generative}. 


Some Imitation Learning (IL) methods have been proposed to overcome the issues, such as Generative Adversarial Imitation Learning (GAIL) \cite{ho2016generative}, Guided Cost Learning (GCL) \cite{finn2016guided}, and Adversarial Inverse Reinforcement Learning (AIRL) \cite{fu2017learning}. Among them, AIRL is capable of learning both a policy and a reward function with better performance. In the original work, AIRL is verified with only simple static environments such as maze navigation of a 2D point mass and a quadrupedal ant running task in OpenAI Gyms. There is no interaction with dynamic objects in the tasks. 

However, driving environments of autonomous vehicles are much more complicated with diverse situations. The interactive behaviors between vehicles have high uncertainties and are difficult to predict and model. In interactive scenarios, decision-making in lane changing tasks are particularly challenging for an autonomous agent. Firstly, the driving task involves complicated objectives. The goal is not just to reach the target lane but also to be safe, efficient, and comfort to the drivers. In contrast, the objectives in previous works~\cite{finn2016guided}\cite{fu2017learning} are simple, e.g. reaching a goal position or velocity. 
Secondly, different from the stationary environments in the OpenAI Gym, driving environments are dynamically changing. Diverse driving scenarios are created by different intrinsic desires and behaviors of surrounding vehicles, which makes the learning more challenging. 
Thirdly, the behaviors of autonomous agents should be human like and therefore understandable by surrounding human drivers. These place difficulties to explicitly design a reward function.

In this study, we extend AIRL to learn decision-making behaviors of lane-change tasks in a simulated environment, where it is highly interactive between the learning agent and surrounding vehicles. The contributions of our work are threefold: 
$1)$ we augment AIRL by concatenating semantic reward terms in the learning framework to improve and stabilize its performance;
$2)$ we adapt augmented AIRL to the more practical and challenging task of lane change in autonomous driving and demonstrate its superior performance over baseline methods;
$3)$ we propose a system design with semantically meaningful actions for decision-making tasks and four representative metrics for evaluation. Besides reaching target lanes, the learnt behaviors are also human-like, i.e. similar to experts' behaviors as shown in the result graphs.

The remainder of the paper is organized as follows. A literature review is presented in Section~\ref{sec:rw}. Section~\ref{sec:Me} introduces the methodology, followed by the description of the decision-making task in Section~\ref{sec:dm} and experiment in Section~\ref{sec:Ex}. Conclusion and discussions are given in the last section.





\section{Related Work} \label{sec:rw}

Some state-of-the-art works see great potential in dealing with the inherent limitations in Imitation Learning methods. Generative Adversarial Imitation Learning (GAIL)~\cite{ho2016generative} combines imitation learning and generative adversarial network to learn a policy against a discriminator that tries to distinguish policy actions from expert actions. Guided Cost Learning (GCL)~\cite{finn2016guided} adopts importance sampling technique to relieve the difficulty in estimating the partition function in the maximum entropy IRL formulation. \cite{finn2016connection} provide a link between GAIL and GCL with a theoretical proof. GAIL with a properly reparameterized discriminator is mathematically equivalent to GCL with some tweaks in importance sampling. 

Adversarial Inverse Reinforcement Learning (AIRL)~\cite{fu2017learning} is constructed on the basis of GAIL and GCL. In contrast to GAIL, AIRL learns both the cost function and the policy; and to GCL, it performs the learning in an adversarial way. The applications of these algorithms are mostly verified with control tasks in OpenAI Gym~\cite{finn2016guided}\cite{ho2016generative}\cite{fu2017learning}. AIRL has shown more robust and superior performance than GAIL and GCL methods, especially when testing environments are shifted from the environment where the expert demonstrates. 


In autonomous driving, many recent studies have applied RL to the vehicle control task~\cite{lillicrap2015continuous}\cite{kuderer2015learning}\cite{wang2019continuous}. Some studies began to apply RL to decision making~\cite{shalev2016safe}\cite{liu2019learning}\cite{wang2018reinforcement}. Only a few studies have applied adversarial learning to driving. One work~\cite{kuefler2017imitating} applied GAIL to learn the lane-keeping task in a driving simulator, with a recurrent neural network as the policy network. The task is relatively simple as lane keeping requires less interactions with road users. To our best knowledge, no prior work has applied AIRL to practical tasks in autonomous driving that needs to deal with interactions with surrounding vehicles. Our work explores its feasibility for handling the challenging decision-making task. 

\section{Methodology} \label{sec:Me}
In this section, we briefly introduce AIRL, then focus on our augmentation of AIRL. 

\subsection{Adversarial Inverse Reinforcement Learning}
AIRL is directly related with maximum entropy IRL~\cite{ziebart2010modeling} and Guided Cost Learning~\cite{finn2016guided}. It uses a special form of the discriminator different from that used in GAIL, and recovers a cost function and a policy simultaneously as that in GCL but in an adversarial way.

The critical element in AIRL is the form of the discriminator. Assuming $\tau=(s_0, a_0, .., s_T, a_T)$ is a behavior sequence from the demonstrated data, and $c_{\theta}(\tau)=\sum_t c_{\theta}(\mathbf{s}_t,\mathbf{a}_t)$  is the unknown cost function parameterized by $\theta$. Let $p(\tau)$ denote the true distribution of the demonstration, and $q(\tau)$ is the generator's density. Based on the mathematical proof in \cite{finn2016connection}, a form of discriminator can be designed as in (\ref{eq7}) where $p(\tau)$ is estimated by the Maximum entropy  IRL distribution. 
\begin{equation} \label{eq7}
D_{\theta}(\tau) = \frac{\frac{1}{Z}\exp(-c_{\theta}(\tau))}{\frac{1}{Z}\exp(-c_{\theta}(\tau))+q(\tau)}
\end{equation}

The optimal solution of this form of discriminator is independent of the generator, which improves the training stability that we will show later in our training results. 

The use of the full trajectory in the cost calculation could result in high variance~\cite{fu2017learning}. Instead, a revised version of the discriminator based on individual state-action pairs is used to remedy this issue. The discriminator can be changed to the following form as in (\ref{eq8}) where $q(\tau)$ is equivalently replaced with $\pi(a|s)$,
\begin{equation} \label{eq8}
D_{\theta}(s,a) = \frac{\exp(f_{\theta}(s,a))}{\exp(f_{\theta}(s,a))+\pi(a|s)}
\end{equation}
$f_{\theta}(s,a)$ is a helper network parameterized by $\theta$.

\subsection{Augmented Adversarial Inverse Reinforcement Learning}
In AIRL, the reward function used for updating the generator is formulated purely with a discriminator in the view of the adversarial learning theory. The agent needs to conduct extensive exploration to understand the rewarding  mechanism  embedded  in  the  demonstrated  behaviors. If we can apply some domain knowledge and assist the learning procedure with some semantic reward signals, it should provide the agent informative guidance and help it learn faster. Based on this insight, we augment the AIRL framework by incorporating some intuitive semantic reward terms. 

\subsubsection{Objectives of Discriminator and Generator}
The discriminator $D_{\theta}(s,a)$ takes state-action pair as input and distinguishes whether it is from expert or generator. $\theta$ represents the network parameters. The objective of the discriminator is to minimize cross-entropy loss between expert data and generated samples as~(\ref{eq9}). 
\begin{equation} \label{eq9}
\begin{split}
\mathcal{L}(D) = & \mathbb{E}_{(s, a)\sim p}[-\log D(s,a)] \\
&+ \mathbb{E}_{(s, a)\sim G}[-\log(1 - D(s,a))]
\end{split}
\end{equation}

The generator's task is to produce samples to fool the discriminator, i.e. to minimize its log probability of being classified as generated samples. This signal alone is not enough to train the generator when the discriminator quickly learns to distinguish between generated data and expert data. Therefore, we add another term, the discriminator's confusion~\cite{finn2016connection}, to the generator's loss.
This way, the generator's objective is to minimize
\begin{equation} \label{eq99}
\mathcal{L}(G) = \mathbb{E}_{(s, a)\sim G}[\log(1 - D(s,a)) - \log D(s,a)]
\end{equation}


\subsubsection{Semantic Reward}
The semantic reward is a way to embed domain knowledge as reward signals. It is referred as semantic because the rewards are semantically meaningful, sparse, and straightforward to craft. Therefore, it is easy to generalize to other decision-making problems. It helps solving the discriminator-generator imbalanced problem, although there are methods proposed to solve this well-known issue, e.g. Wassertein GAN~\cite{gulrajani2017improved} and its imitation learning application~\cite{xiao2019wasserstein}. In this paper, we show that adding semantic rewards is an effective way to improve the performance and speed up the learning process.

In our case, the goal is to successfully complete the lane change efficiently and not to collide with other objects nor invade their safety margins. Therefore, the semantic rewards $r_{sem}$ is composed as $r_{sem}=\{r_{suc}, r_{col}, r_{mgn}, r_{mov}\}$, with their respective trainable weights $w_{sem}=\{w_{suc}, w_{col}, w_{mgn}, w_{mov}\}$. Their state-dependent values are defined as follows, $r_{suc}=15~\mathrm{if}\ \mathrm{success}$, $r_{col}=-30~\mathrm{if}\ \mathrm{crash}$, $r_{mgn}=-1~\mathrm{if}\ \mathrm{invade}\ \mathrm{safety}\ \mathrm{ margin}$,  and $r_{mov}=0.3~\mathrm{if}\ \mathrm{lateral}\ \mathrm{move}$. The weights $w_{sem}$ are initialized to be a vector of 1s. The total semantic reward is the dot product of $r_{sem}$ and $w_{sem}$.

The reward values are assigned based on the principle that the total returns of each semantic rewards in an episode are in similar magnitude, as well as that a collision should incur a large penalty. As the reward weights are trained, the optimization is not sensitive to the exact reward values we assign.

\subsection{Augmentation and Optimization}

One important question is how to incorporate the semantic reward terms to the learning procedure. The most common and intuitive way is to add $r_{sem}(s,a)$ directly to the generator reward function Eq. \ref{eq12}, which is known as reward shaping~\cite{ng1999policy}. Some work~\cite{perot2017end} might show effectiveness by doing it this way, but one obvious drawback is that the value assigned to the semantic reward term is subjective and case sensitive. More importantly, our experiments show that it only helps marginally in our case. One reason could be that the semantic reward term only affects the updates in the generator and leaves the discriminator untouched, which fails to synchronize the performance of the discriminator and the generator.
 \begin{equation}
 \label{eq12}
\begin{split} 
 R(s,a) &= \log(D_{\theta}(s,a)) - \log (1-D_{\theta}(s,a))\\
        &= f_{\theta}(s,a) - \log \pi(a|s)
\end{split}
\end{equation}

We add $r_{sem}(s,a)$ to the output of the discriminator helper network, $f_{\theta}(s,a)$, it effectively adds the same term to the generator reward $R(s,a)$, but it consistently changes the discriminator as well. This approach can also be interpreted as reparameterization of the discriminator network to better capture the reward mechanism.



\begin{figure}[b]
    \includegraphics[width=0.40\textwidth]{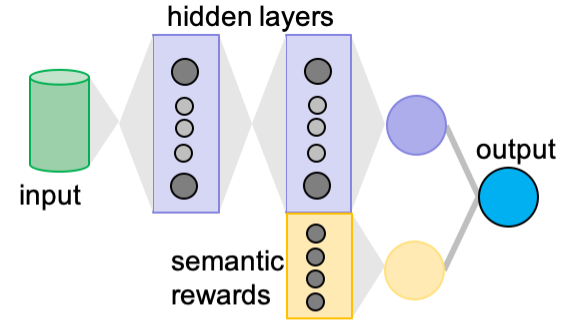}
    \caption{The architecture for the discriminator helper function $f_{\theta}(s,a)$. The semantic reward is added to the last layer.}
    \label{fig:d}
\end{figure}

Fig.~\ref{fig:d} illustrates the architecture of augmented discriminator helper function. The final discriminator can be expressed as follows.
\begin{equation} \label{eqNewRe}
D_{\theta}(s,a) = \frac{\exp(f_{\theta}(s,a)+r_{sem}*w_{sem})}{\exp(f_{\theta}(s,a)+r_{sem}*w_{sem})+\pi(a|s)}
\end{equation}

\begin{figure}[b]
    \includegraphics[width=0.45\textwidth]{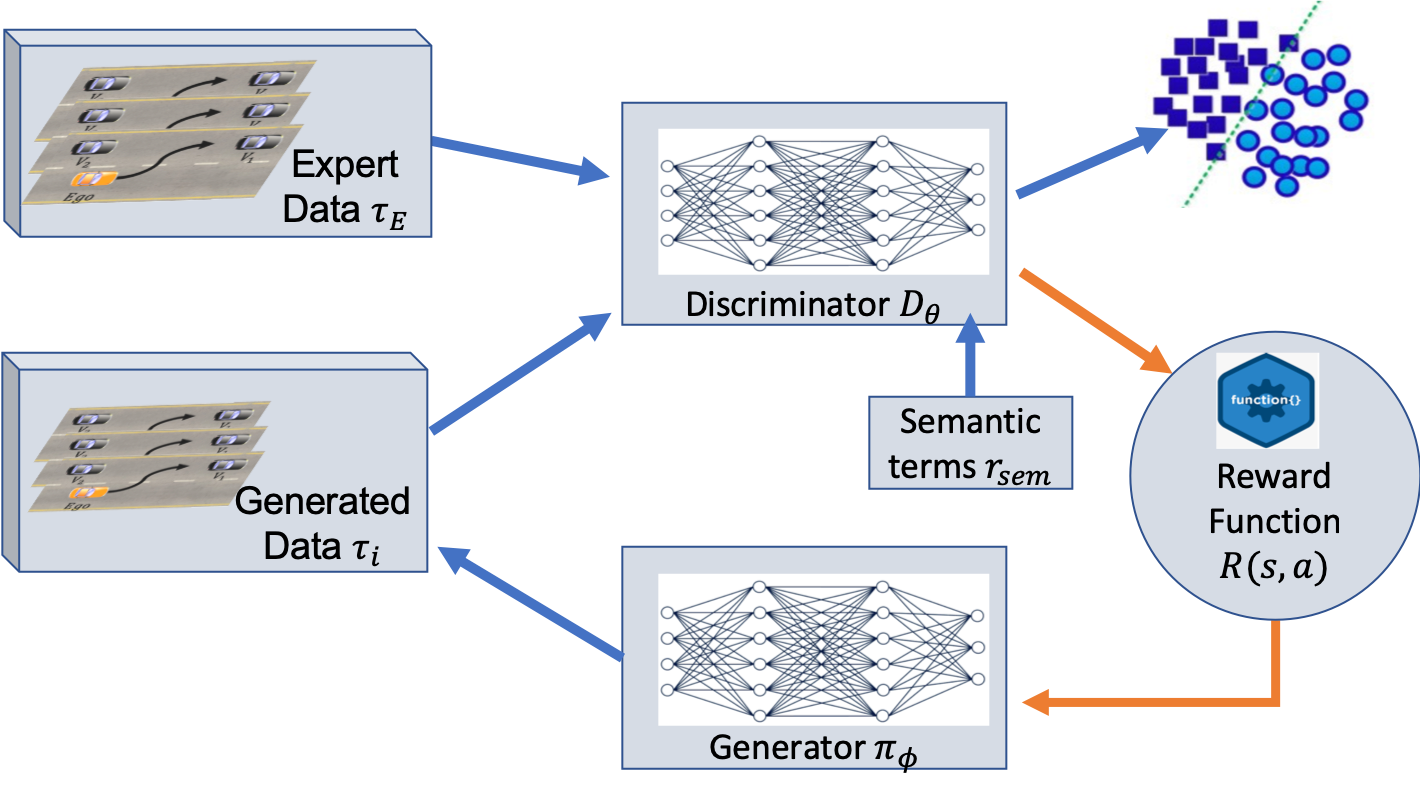}
    \caption{Augmented AIRL learning framework. The weights of discriminator consist the reward function for policy update.}
    \label{fig:o}
\end{figure}

The overall learning framework is illustrated in Fig.~\ref{fig:o}. The core of the training procedure of the augmented AIRL method is in Algorithm 1. The optimization of the discriminator and generator is similar to GCL where rewards learning interleaves with the policy optimization procedure. The difference is the updating method is done in an adversarial way. 

The optimization of discriminator is based on (\ref{eq9}) as a binary logistic regression problem. The updated discriminator augmented with the semantic rewards is fed to the generator objective (\ref{eq99}) to update the policy. For the policy optimization, in GCL it is based on guided policy search~\cite{levine2013guided}, a model-based method where the environment model is learned with a time-varying linear dynamic model~\cite{finn2016guided}. As the dynamic environment is too complicated to learn for our driving task, we instead use a model-free policy optimization method, Trust Region Policy Optimization (TRPO)~\cite{schulman2015trust}, for the policy update.

\begin{algorithm}[t!]
\caption{Augmented AIRL}
\label{alg:AIRL}
\setlength{\lineskip}{1pt}
\begin{algorithmic}[1]
\State Collect expert lane-change sequences $\tau_E$
\State Augment discriminator $D_{\theta}$ with semantic rewards $r_{sem}$
\State Initialize $D_{\theta}$ including weights for $r_{sem}$
\State Initialize decision-making policy $\pi_{\phi}$
\For {iteration $i=1$ to $N$} 
\State Collect decision-making sequences $\tau_i=(s_0, a_, .., s_{T-1}, a_{T-1})$ with current policy $\pi_{\phi i}$
\State Train  $D_{\theta}$ based on (\ref{eq9}) with data drawn equally from $\tau_E$ and $\tau_i$ for $m$ iterations
\State Train policy $\pi_{\phi}$ based on (2) w.r.t. augmented reward in (\ref{eqNewRe}) using TRPO for $n$ iterations
\EndFor
\end{algorithmic}
\end{algorithm}

\section{System Design for Decision-making Task} \label{sec:dm}

\begin{figure}[b]
  \centering
    \includegraphics[width=\linewidth]{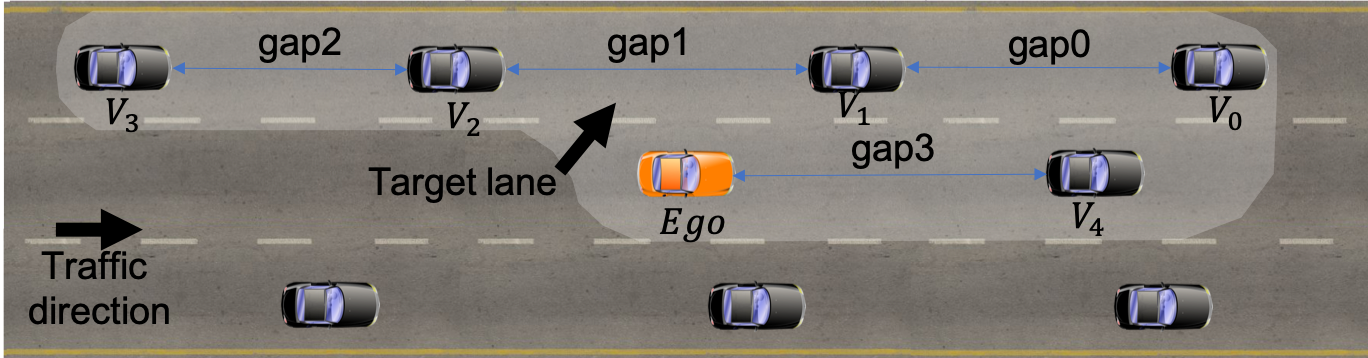}
    \caption{Illustration of decision making in a lane-change scenario. Longitudinal decision corresponds to the target gap selection from 0 to 3. Lateral decision is whether to execute the lateral movement at the current step.}
    \label{fig2}
\end{figure}

To learn decision-making strategies from demonstrations, the design of action space and state space plays a critical role. In order to imitate human-like driving behaviors, we design the action space in comparable to how human makes decisions. Instead of acceleration and steering, human express decisions in a high level, such as merge to the gap ahead of me (e.g. gap 0 in Fig.~\ref{fig2}) or wait for the next gap (e.g. gap 2 in Fig.~\ref{fig2}). With this consideration, we design the action space as in Table~\ref{table:a}. In this way, the learnt decisions are semantically meaningful, and can be potentially used to guide the motion planning in related studies.

\subsection{Action Space}\label{subsec:Action}
To reproduce the diverse human driving decisions in lane change scenarios, we separate the decision into longitudinal and lateral directions. The longitudinal decision corresponds to the target gap selection, e.g. gap 0, gap 1, gap 2 or gap 3 in Fig.~\ref{fig2}. These decisions allow the ego vehicle to adjust its longitudinal position and speed relative to the leading vehicle of the selected gap, and therefore prepare for committing lane changes. The lateral decision is binary, i.e. either stay on its current lane or move laterally toward the target lane immediately. Combining both longitudinal and lateral directions and ignoring the unreasonable combinations, we keep five actions in action space, as shown in Table~\ref{table:a}. 

\begin{table}[b]
\caption{Action space design}
\begin{center}
\begin{tabular}{|c|l|}
\hline
\multirow{1}{*}{\textbf{Action pairs}} & \multicolumn{1}{l|}{\textbf{Semantic meaning}} \\
\hline
$a_{0} \ (0,0)$ & Move ahead to gap 0 and keep current lane\\ 
$a_{1} \ (1,0)$ & Move alongside with gap 1 and keep current lane\\ 
$a_{2} \ (1,1)$ & Move alongside with gap 1 and change to target lane\\ 
$a_{3} \ (2,0)$ & Move behind to gap 2 and keep current lane \\ 
$a_{4} \ (3,0)$ & Keep lane with gap 3 and follow current leading vehicle  \\ 
\hline
\end{tabular}
\label{table:a}
\end{center}
\end{table}

The designed actions are meaningful and explainable. It can reflect the diverse interactions among the learning agent and the relevant vehicles, and also enables the agent to perform more complicated human-like behaviors. One interesting observation is that the agent learns to perform the behavior of "abort", which is often missing in many decision-making studies. When the vehicle agent detects itself in a risky lane-change situation, it can choose to abort the lane change and return back to its original lane. This is an appealing feature that not many learning-based decision-making studies have demonstrated.

 The decisions are executed by a group of low-level controllers which output the control commands, e.g. steering, acceleration. This system design is consistent with industrial pipelines where low-level controllers are well developed and equipped. In our simulation, we use a PID controller \cite{aastrom1995pid} for lateral control where the center of the target lane is set as the reference, and a sliding-mode controller \cite{edwards1998sliding} for longitudinal control where a desired time gap to the leading vehicle is set as the reference. Advance methods like model predictive control could be used, but we find performances are satisfiable with current settings.

\subsection{State Space}
In daily driving,  we usually only pay attention to certain parts of the scene related to driving tasks, rather than to all the information in the observable field of view. This subconscious act helps us lower the burden of scene understanding and focus more on the relevant information. Similar ideas should be applied to the learning agent. Instead of adopting all the available information from sensors, only relevant observations are provided. Hence, the state space in our study only includes features from relevant vehicles, i.e. vehicles $V_{0}, V_{1}, V_{2}, V_{3}, V_{4}$ forming gap 0 to gap 3, as shown in Fig.~\ref{fig2}. In this study, we refer to observations and states interchangeably, and do not consider hidden states.  

For each relevant vehicle, we gather vehicle kinematics that we can derive from on-board sensors, such as vehicle speed, acceleration, position, lane id, and vehicle id. For the ego vehicle, we also gather similar information plus a target lane id. In total, there are 44 features.


\section{Experiment} \label{sec:Ex}

In this study, we use simulation for testing and evaluation. First, it is easier to generate diverse scenarios to cover dynamic and uncertain interactions in simulation. Second, the training of the policy and the learned reward function with AIRL needs an interacting environment to gather new data and bootstrap the learning direction of the agent. Third, the simulator provides a safe environment to test the performance and explore limitations. The learned model can be refined easily with real-world driving data as we design the input state and output actions with much practical consideration. 

The baseline methods can be classified into two categories. One is Imitation Learning methods such as AIRL and GAIL. The other is Reinforcement Learning methods such as pure RL (TRPO) and a combination of Behavior Cloning and RL (BC+TRPO).  

Expert demonstrations are generated by a well-defined rule-based system in the simulator which requires full access to the behavioral models of surrounding vehicles. Given the entire knowledge of the simulated environment, it is possible to perceive the interactions between vehicles, thereby generating reasonable expert behaviors as demonstrated data. The data used for training can be easily replaced with real-world data if rich driving scenarios and behaviors are available. The expert behaviors and qualitative results of augmented AIRL agents are illustrated in the video, \url{https://youtu.be/tEnQhtFk3DM}. 

\subsection{Simulation Environment}
\begin{figure}[b]
    \includegraphics[width=\linewidth]{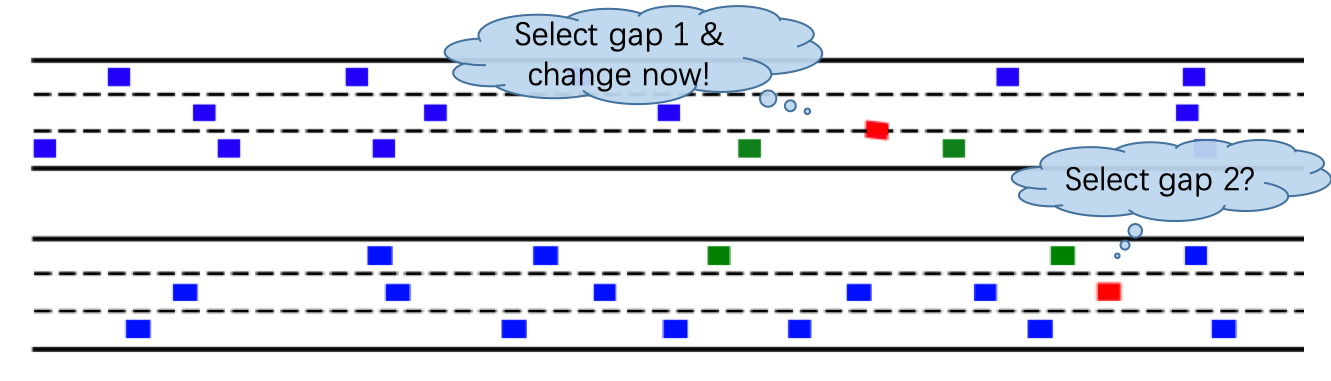}
    \caption{Demonstrations of decision-making strategies for lane change behavior. The upper scene shows that the ego vehicle decides to commit lane change to gap 1. The lower scene shows that it decides to wait to merge to gap 2.}
    \label{fig3}
\end{figure}

The simulation environment is a highway segment with three lanes on each direction. To be as realistic as possible, we develop many features for dynamic traffics and diverse interactions. Each vehicle has a different initial speed, a desired speed limit, and a desired headway to leading vehicles. To generate diverse traffic scenes, the time gap between vehicles and traffic densities between lanes are also randomized. For interaction behaviours during a lane change, surrounding vehicles can give way to or overtake the ego vehicle based on the randomized trigger conditions. Aborting behavior happens when the lateral position or lateral speed of the ego vehicle is in a certain risk regions of the interacting vehicle. With the variations in the vehicle behaviors and traffic densities, we are able to create a bunch of diverse driving demonstrations for training. An illustration of simulated decision-making scenarios is shown in Fig.~\ref{fig3}.

\subsection{Comparison Methods and Evaluation Metrics}
As AIRL and GAIL are bench-marking Imitation Learning methods, and AugAIRL is closely related to these two, the comparisons are made among the three methods. Since they are trained with the same expert demonstrations, the performance of expert is considered as guidelines. In addition, since RL based methods are also promising if a reward function is available, we also compare AugAIRL with a policy gradient RL method TRPO, and a method integrating Behavior Cloning and TRPO (BC+TRPO). The reward function in these two methods are defined according to the expert driving rules in the simulator, i.e. with consideration in safety, efficiency, and comfort.  


The selection of proper evaluation metrics is important to assess the performance of the algorithms. Simply using the accumulated reward and episode length, as in \cite{finn2016guided} and \cite{fu2017learning}, is not enough for evaluating complicated lane-change behaviors. In our study, we propose to use four metrics: 
$1)$ the success ratio, which calculates with the number of successful lane-change cases over the total number of experiments;
$2)$ the number of decision-making steps, which is consumed by a lane-change process and counted from the moment the lane change command is issued to the time when the lane-change process finishes;
$3)$ the number of changing-lane steps, which is counted from the start of lateral movement to the completion;
$4)$ the total reward, which is accumulated during the lane-changing procedure. The reward value is derived from a carefully crafted function as used in RL baselines. In comparison to semantic rewards, it is designed particularly for these lane change scenarios, therefore does not generalize easily. It reflects the complicated rules and logic of safety, efficiency, and comfort. Notice that as the true reward function for driving tasks is unknown, such a reward value is used as a consistent indicator for evaluating the performance across different methods.

\subsection{Training Results}

\begin{figure}[t]
\centering
\subfloat[\textbf{Total reward}]{\includegraphics[width=.44\linewidth]{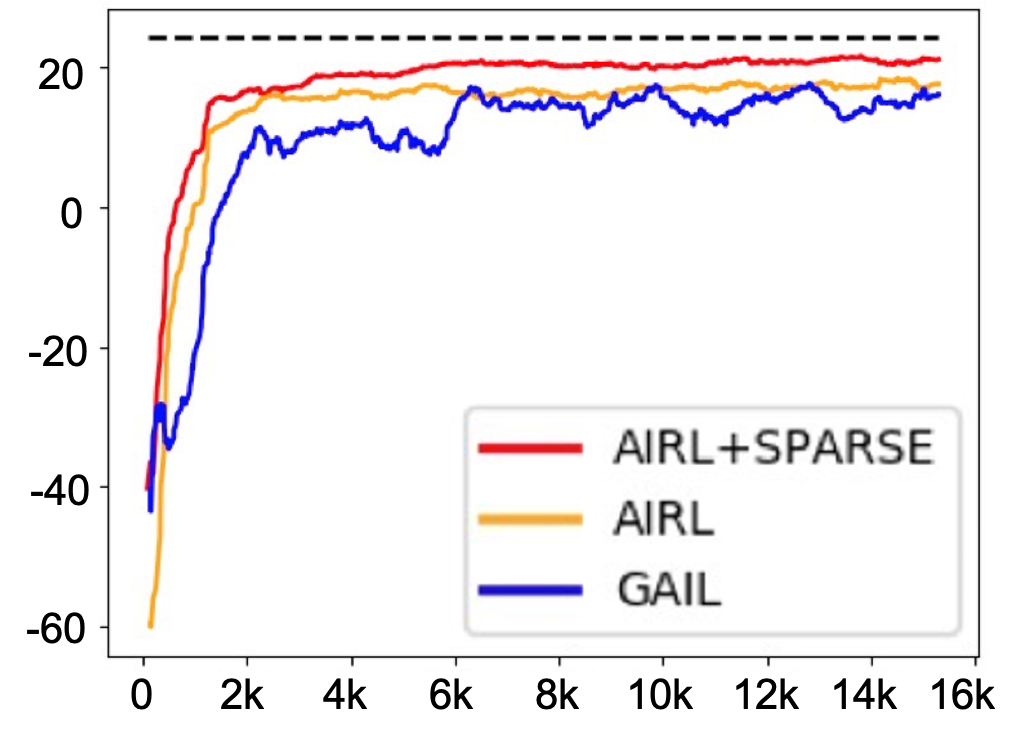}\label{fig:tr}}\quad
  \subfloat[\textbf{Success ratio}]{\includegraphics[width=.44\linewidth]{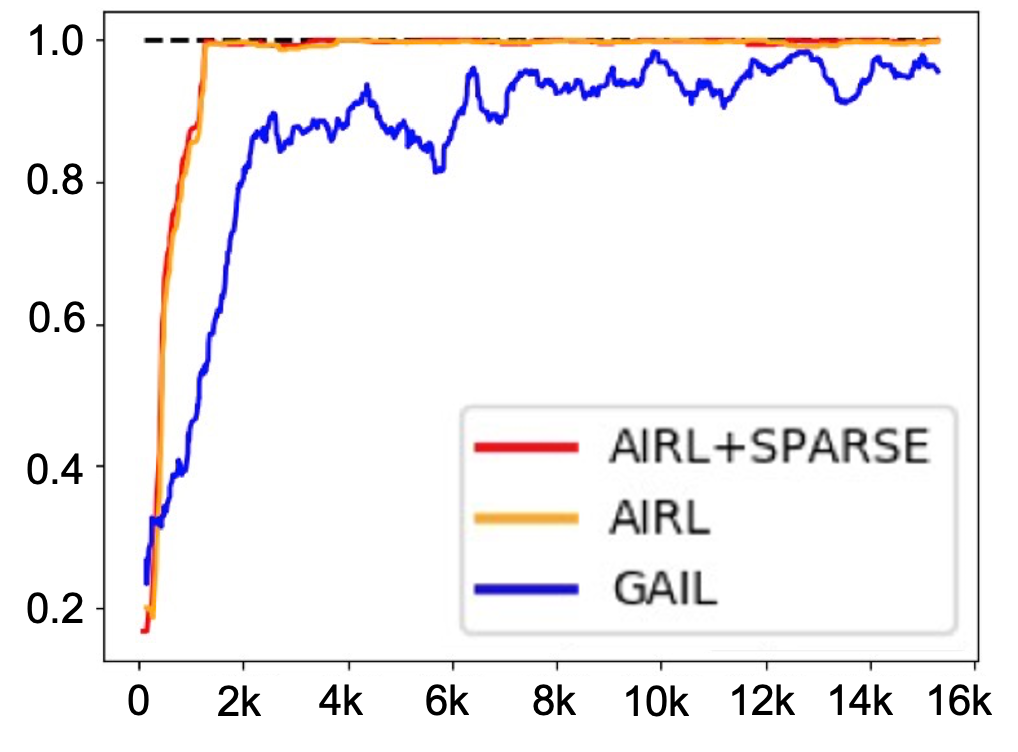}\label{fig:sr}}\\
  \subfloat[\textbf{Decision steps}]{\includegraphics[width=.44\linewidth]{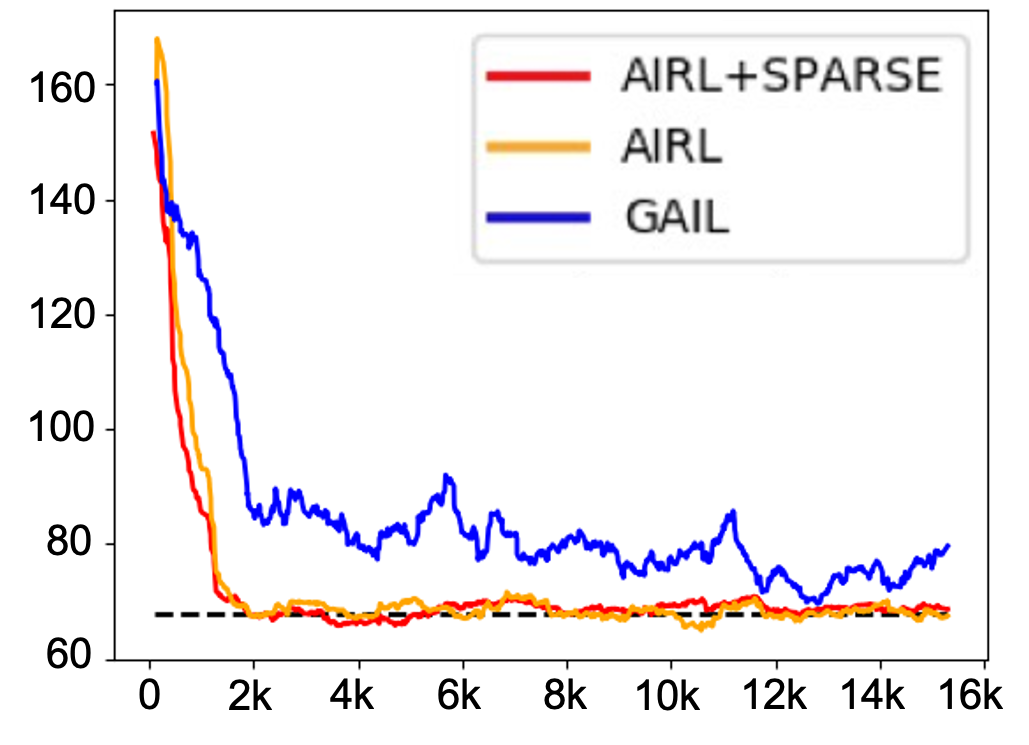}\label{fig:ds}}\quad
  \subfloat[\textbf{Changing steps}]{\includegraphics[width=.44\linewidth]{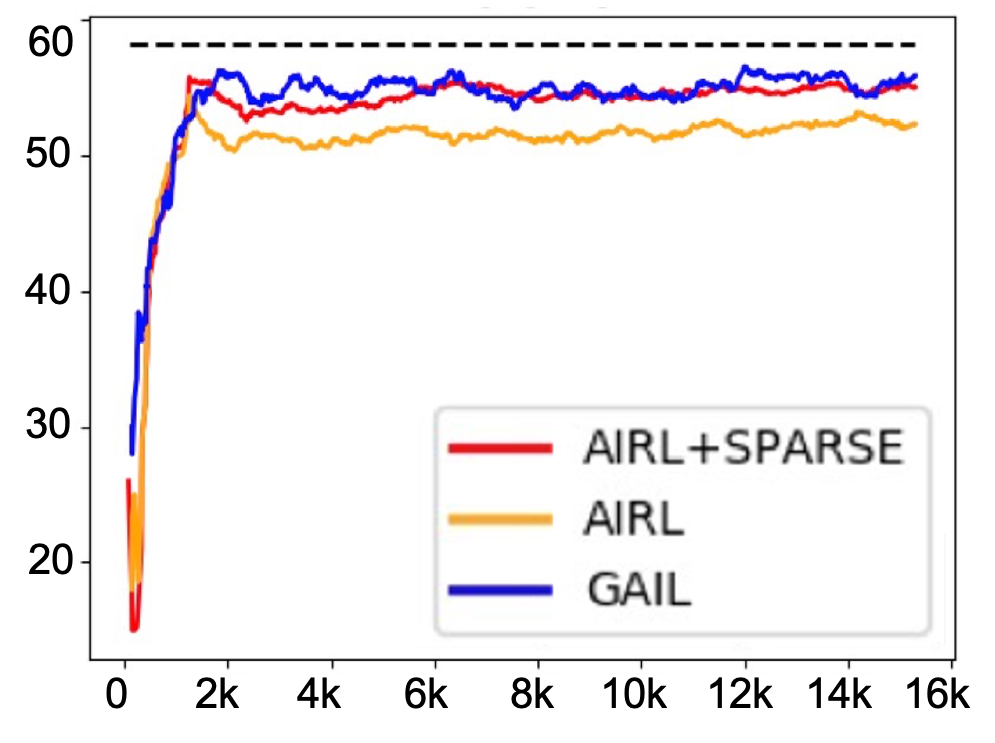}\label{fig:cs}}
  
  \caption{Training results of GAIL (blue), AIRL (orange) and AugAIRL (red), as well as expert values (black) of four metrics over 15,000 iterations (x-axis). AIRL based methods demonstrate better performance and converge faster than GAIL. AugAIRL shows the best overall performance.}
  \label{fig:training}
\end{figure}

Fig.~\ref{fig:training} depicts the training curves of the four metrics for the three imitation learning based methods, plus the expert values. Table~\ref{table1} gives the quantitative results of the mean and variance of the model saved at the last iteration in all the algorithms. It can be seen that AugAIRL has the best results in every metric. For total reward and success ratio, high value is preferred. For decision and changing steps, it should be close to expert, which indicates the human-like behaviors. In general, better performance has closer mean value to the experts' and smaller variance.

\begin{table*}[t] 
\caption{Training results in four evaluation metrics from the trained models with 15k iterations}
\centering
\begin{tabular}{|c|p{0.05\textwidth}|p{0.05\textwidth}|p{0.05\textwidth}|p{0.05\textwidth}|p{0.05\textwidth}|p{0.05\textwidth}|p{0.05\textwidth}|p{0.05\textwidth}|p{0.05\textwidth}|p{0.05\textwidth}|}
\hline
\multirow{2}{*}{\textbf{Algorithms}} & \multicolumn{2}{c|}{\textbf{Total Reward}} & \multicolumn{2}{c|}{\textbf{Success Ratio}} & \multicolumn{2}{c|}{\textbf{Decision Steps}} & \multicolumn{2}{c|}{\textbf{Changing Steps}} & \multicolumn{2}{c|}{\textbf{Discrim. Loss}} \\ \cline{2-11} 
                                     & \textbf{Mean}        & \textbf{Std}        & \textbf{Mean}         & \textbf{Std}        & \textbf{Mean}         & \textbf{Std}         & \textbf{Mean}         & \textbf{Std}         & \textbf{Mean}       & \textbf{Std}      \\ \hline
\textbf{TRPO}                      & 14.89                & 7.93                  & 0.46                  & 0.17                  & 129.78                 & 15.19                   & 54.86                 & 9.30                   & --                  & --                \\
\hline
\textbf{BC+TRPO}                    & 15.11                & 6.54                  & 0.41                  & 0.21                  & 122.38                 & 13.48                   & 56.29                 & 7.59                   & --                  & --                \\
\hline

\textbf{GAIL}                        & 15.67                & 5.78                & 0.955                 & 0.071               & 77.44                 & 12.82                & 55.57                 & 3.6                  & --                  & --                \\ \hline
\textbf{AIRL}                        & 17.58                & 4.58                & \textbf{0.99}                  & \textbf{0.01}                & \textbf{67.54}                 & 8.41                 & 52.53                 & 3.56                 & 0.694               & \textbf{0.032}             \\ \hline
\textbf{AugAIRL}                 & \textbf{22.12}                & \textbf{2.81}                & \textbf{1}                  & \textbf{0.01}                &  \textbf{65.58}                 & \textbf{5.66}                 & \textbf{58.28}                & \textbf{3.36}                & \textbf{0.625}               & \textbf{0.035}             \\ \hline
\textbf{Expert}                      & 24.21                & --                  & 1.00                  & --                  & 67.74                 & --                   & 58.18                 & --                   & --                  & --                \\
\hline
\end{tabular}
\label{table1}
\end{table*}

As shown in Fig. \ref{fig:training}, the reward curve of the AugAIRL is the highest compared with AIRL and GAIL, and most closely to the expert's value. Success ratio and decision steps show good performance in AugAIRL and AIRL, while changing steps in AugAIRL and GAIL have comparable performance. The overall results show the superiority of AugAIRL in all the four metrics that closely match with the expert performance. It is also noticeable that the AIRL-based methods are more stable than GAIL, proving the effectiveness of the special form of the discriminator.

More interesting results can be obtained from exploring the curves of AIRLs. From the successful ratio curves in Fig.~\ref{fig:sr}, the agent completes lane changes with a high successful rate, approaching to the expert performance of 1.0. It is particular promising since the agent learns the decision-making strategy from scratch. 

In Fig.~\ref{fig:ds} and Fig.~\ref{fig:cs}, the values of the decision-making steps and lane-changing steps of AugAIRL indicate that the lane change behavior at convergence stays quite stable and the maneuvering time required is quite similar to that of the expert. Additionally, from the shape of the these curves, we can infer that at the beginning of the training, the agent is quite conservative and hesitates to commit lane change as the decision steps are high (Fig.~\ref{fig:ds}) and the actual lane-changing steps are low (Fig.~\ref{fig:cs}), i.e. no lateral movement trails. As training goes on, the agent actively explores and finally learns to imitate the demonstrated behaviors.

We omitted the training curves of the RL based methods (TRPO and BC+TRPO) due to space limitation but summarized their mean and variance at the lastly saved model as in Table~\ref{table1}, together with the means and variances of other methods. We can confirm that the mean values of almost all the metrics in AugAIRL are the closest to the experts', and the variances are also the lowest. 

\begin{figure}[t]
\centering
  \subfloat[\textbf{Total reward}]{\includegraphics[width=.44\linewidth]{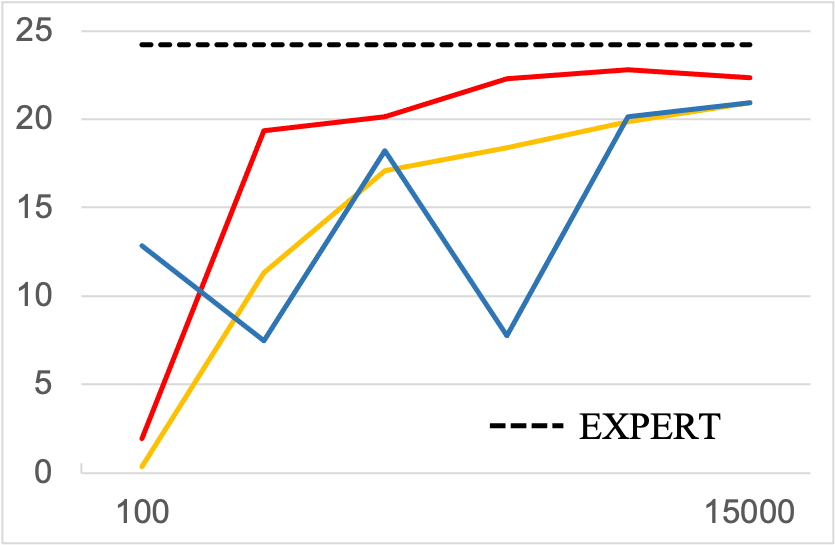}}\quad
  \subfloat[\textbf{Success ratio}]{\includegraphics[width=.44\linewidth]{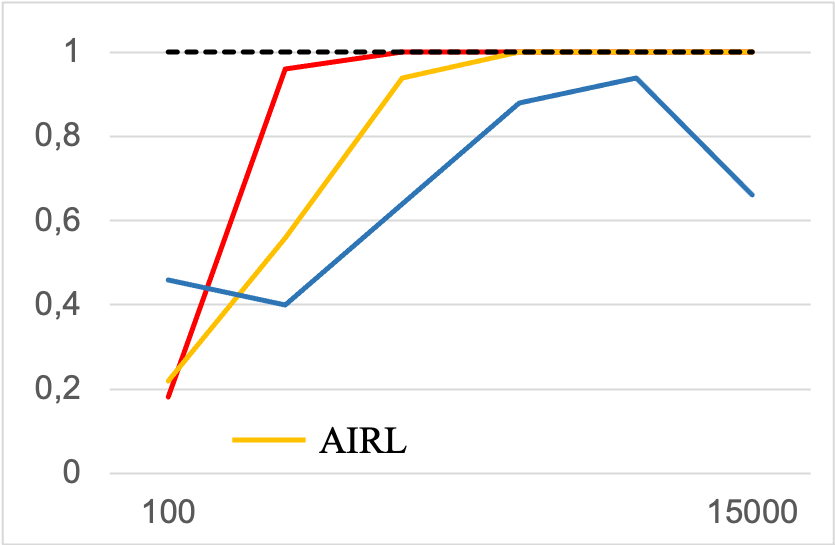}}\\
  \subfloat[\textbf{Decision steps}]{\includegraphics[width=.44\linewidth]{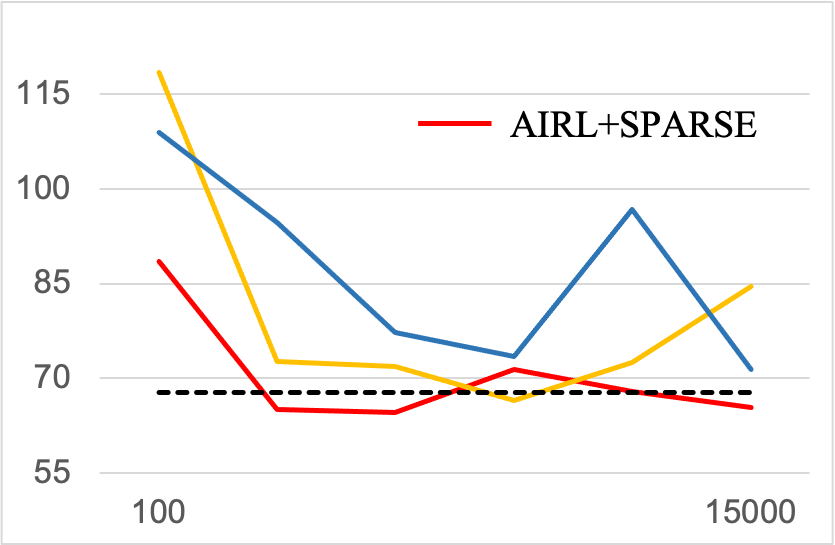}}\quad
  \subfloat[\textbf{Changing steps}]{\includegraphics[width=.44\linewidth]{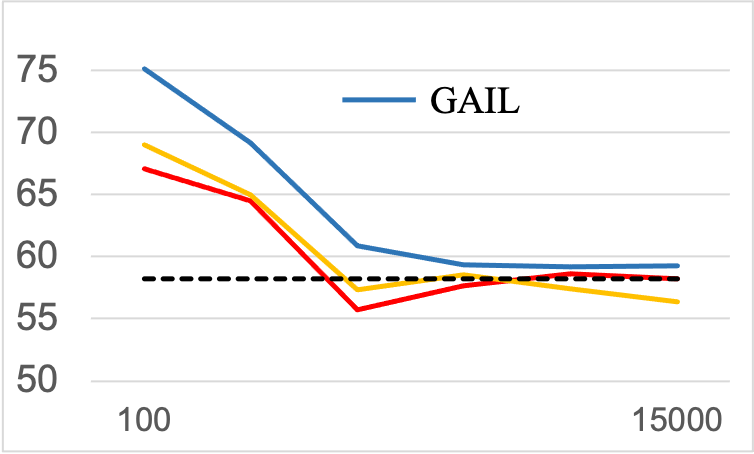}}
  \caption{Testing results of AugAIRL (red) and other Imitation Learning based methods, GAIL (blue) and AIRL (orange), in comparison with expert values (black), for four metrics over 5 saved checkpoints. X-axis is the overall iteration steps and shows where the checkpoints are saved.}
  \label{fig:result}
\end{figure}

\subsection{Validation Results}
\begin{figure}[t]
\centering
  \subfloat[\textbf{Total reward}]{\includegraphics[width=.44\linewidth]{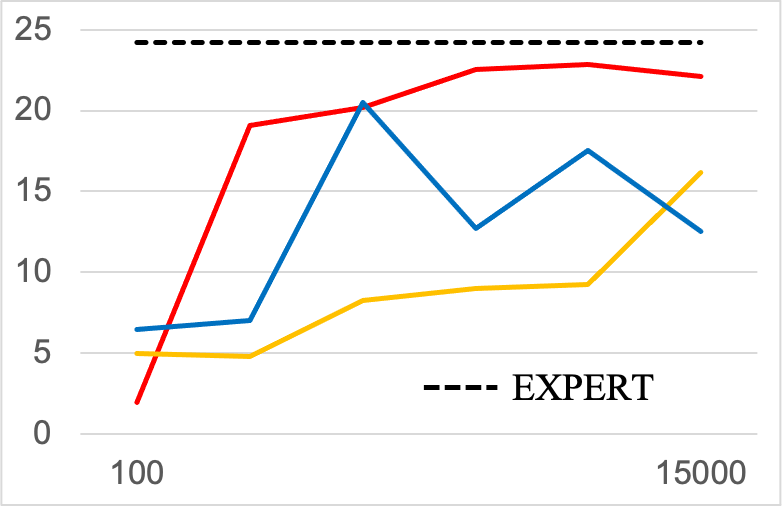}}\quad
  \subfloat[\textbf{Success ratio}]{\includegraphics[width=.44\linewidth]{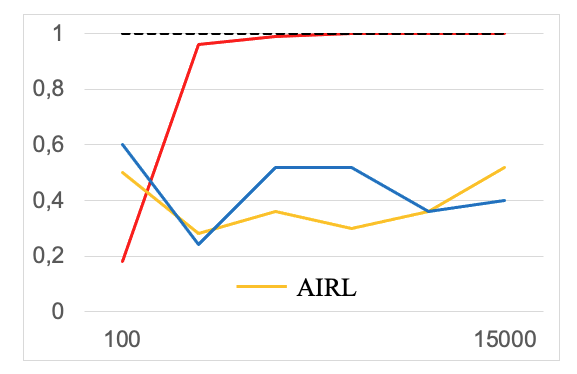}}\\
  \subfloat[\textbf{Decision steps}]{\includegraphics[width=.44\linewidth]{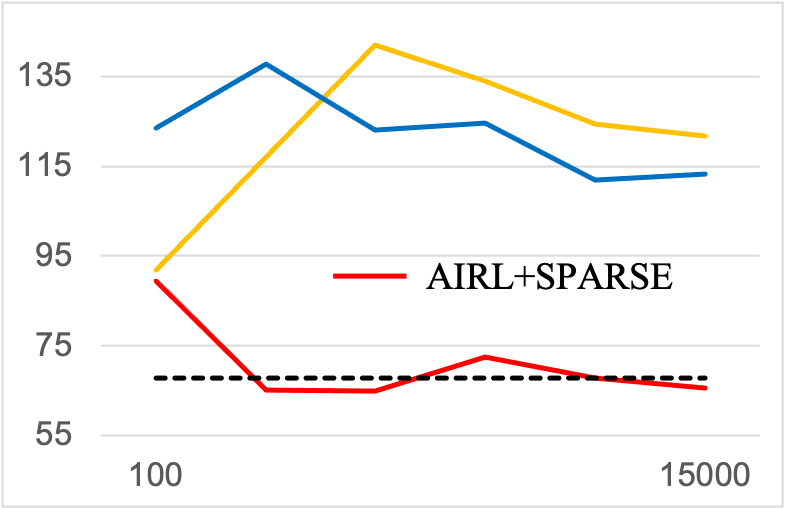}}\quad
  \subfloat[\textbf{Changing steps}]{\includegraphics[width=.44\linewidth]{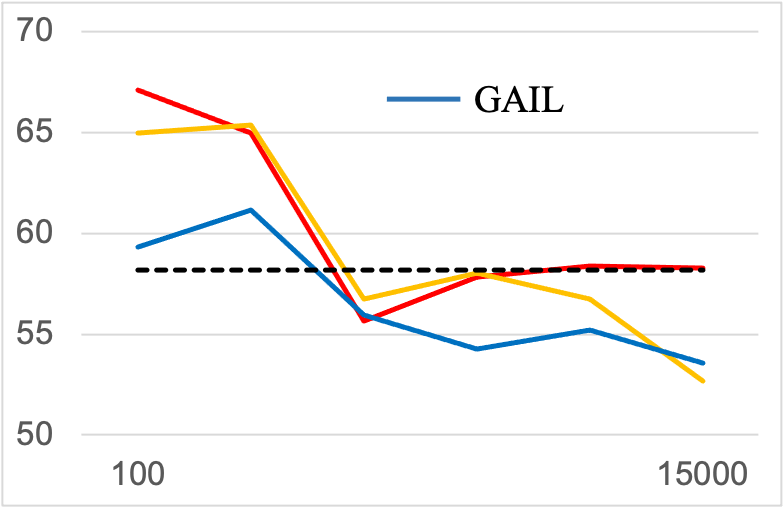}}
  \caption{Testing results of AugAIRL (red) and other RL methods, TRPO (yellow) and BC+TRPO (blue), in comparison with expert values (black), for four metrics over 5 saved checkpoints. X-axis is the overall iteration steps and shows where the checkpoints are saved.}
  \label{fig:reBC}
\end{figure}

We conduct testing on 5 checkpoints saved during training for each method. At each checkpoint, we run 50 episodes and average their metric values. Fig.~\ref{fig:result} and Fig.~\ref{fig:reBC} show the curves in testing with expert data plotted in dashed lines. 

Testing results in both Fig.~\ref{fig:result} and Fig.~\ref{fig:reBC} demonstrate that AugAIRL has the best performance either compared with Imitation learning based method (i.e. AIRL and GAIL) or RL based methods (i.e. TRPO and BC+TRPO).  Its total reward is generally higher, its success ratio approaches 1.0 faster, and its decision-making steps and lane-changing steps are quite equivalent to the expert. The difference between AugAIRL and all the other methods is more significant in testing than in training, and its performance is more stable.

\section{Conclusions and Discussions} \label{sec:con}
In this paper, we augment Adversarial Inverse Reinforcement Learning by incorporating state-dependent semantic reward terms to the discriminator network. As the reward weights are trainable, the network has the flexibility to learn the magnitudes of the rewards. With AugAIRL, we can recover not only the policy but also the reward function that is considered to be potentially useful for adaptations to different environments.

We test our method on the decision-making task of lane changing in autonomous driving, which is challenging as it involves many highly interactive agents. The method is compared with imitation learning based methods (i.e. AIRL and GAIL) and RL based methods (i.e. TRPO and BC+TRPO). Training and testing results in simulation show that our proposed augmented AIRL yields the most satisfactory performance in all of the four metrics (i.e. total reward, success ratio, decision-making steps and lane-changing steps) and is comparable to expert performance. We consider the augmentation as a lightweight engineering process and we believe the approach can be generalized to other imitation learning problems as well.

\bibliographystyle{IEEEtran}
\bibliography{IEEEabrv,example}

\end{document}